\documentclass{article}
\usepackage[english]{babel}

\usepackage{amsthm}
\usepackage{amsmath}
\usepackage[noend]{algpseudocode}
\usepackage{array}
\usepackage{xcolor}
\usepackage[acronym, toc, nonumberlist]{glossaries}
\usepackage{booktabs}
\usepackage{algorithm}
\usepackage{caption}
\usepackage{subcaption}
\usepackage{csvsimple}
\usepackage{siunitx}
\usepackage{makecell}
\usepackage{tikz}
\usepackage{BayesNet}
\usepackage{cleveref}
\usepackage{dblfloatfix}
\usepackage{url}
\theoremstyle{definition}
\newtheorem{definition}{Definition}[]

\theoremstyle{definition}

\DeclareMathOperator*{\maximize}{maximize}
\DeclareMathOperator*{\argmax}{arg\,max}

\newacronym{DTW}{DTW}{Dynamic Time Warping}
\newacronym{GMM}{GMM}{Gaussian Mixture Model}
\newacronym{LCSS}{LCSS}{Longest Common Subsequence}
\newacronym{MVM}{MVM}{Minimal Variance Matching}
\newacronym{CVIs}{CVIs}{Clustering Validity Indices}
\newacronym{GAK}{GAK}{Global Alignment Kernels}
\newacronym{TGAK}{FGAK}{Fast Global Alignment Kernels}
\newacronym{VI}{VI}{Variation of Information }
\newacronym{ETA}{ETA}{Estimated time for accomplishment}
\newacronym{DTW2}{DTW2}{Dynamic Time Warping using L2 norm}
\newacronym{DBA}{DBA}{DTW Barycenter Averaging}
\newacronym{PAM}{PAM}{Partitions Around Medoids}
\newacronym{BN}{BN}{Bayesian Network}
\newacronym{BM}{BM}{Bayesian Multinet}
\newacronym{DBN}{DBN}{Dynamic Bayesian Network}
\newacronym{DBM}{DBM}{Dynamic Bayesian Multinet}
\newacronym{GM}{GM}{Graphical Model}
\newacronym{DAG}{DAG}{Directed Acyclic Graph}
\newacronym{LOCF}{LOCF}{Last Observation Carried Forward}
\newacronym{MCAR}{MCAR}{Missing Completely at Random}
\newacronym{MAR}{MAR}{Missing at Random}
\newacronym{MNAR}{MNAR}{Missing not at Random}
\newacronym{MLE}{MLE}{Maximum Likelihood Estimation}
\newacronym{CPD}{CPD}{Conditional Probability Distribution}
\newacronym{EM}{EM}{Expectation Maximization}
\newacronym{SEM}{SEM}{Structural Expectation-Maximization}
\newacronym{ESS}{ESS}{Expected Sufficient Statistics}
\newacronym{LL}{LL}{Log-Likelihood}
\newacronym{MDL}{MDL}{Minimum Description Length}
\newacronym{BIC}{BIC}{Bayesian Information Criterion}
\newacronym{CkG}{C$\kappa$G}{Consistent $\kappa$-Graph}
\newacronym{BCkG}{BC$\kappa$G}{BFS-Consistent $\kappa$-Graph}
\newacronym{BFS}{BFS}{Breadth-First Search}
\newacronym{TAN}{TAN}{Tree-Augmented Naive}
\newacronym{CML}{CML}{Classification Maximum Likelihood}
\newacronym{ml}{ml}{Maximum Likelihood}
\newacronym{CEM}{CEM}{Classification Expectation Maximization}
\newacronym{HMM}{HMM}{Hidden Markov Model}
\newacronym{CVI}{CVI}{Clustering Validity Indice}

\begin{document}
\title{Imputation in time series}
\author{S. Arcadinho \;\;\; P. Mateus\\[2mm]
IST, ULisbon\\ Instituto de Telecomunica\c{c}\~{o}es}
\date{}

\maketitle

\begin{abstract}
Multivariate time series is a very active topic in the research community and many machine learning tasks are being used in order to extract information from this type of data. However, in real-world problems  data has missing values, which may difficult the application of machine learning techniques to extract information. In this paper we focus on the task of imputation of time series. 
Many imputation methods for time series are based on regression methods. Unfortunately, these methods perform poorly when the variables are categorical. To address this case, we propose a new imputation method based on Expectation Maximization over dynamic Bayesian networks.
The approach is assessed with synthetic and real data, and it outperforms several state-of-the art methods.
\end{abstract}


\section{Introduction}
Nowadays the world is full of digital data, due to the large deployment of sensors, fast internet and more computational power to generate all such that data. This data is might be very useful to extract information and predict events, allowing us to control or profit from them. In order to achieve such goal, we need fast algorithms that are capable of finding features that could bring useful information. However, this is a non-trivial task, as data is very large and usual simple statistics are slow and inaccurate. Thus, 
the term data mining appeared to describe the problem of finding useful information in large data sets by integrating methods from many fields, like machine learning, statistics and database systems, spatial or temporal data analysis, pattern recognition,  image and signal processing.

In recent years many works have been done to use machine learning techniques in order to extract useful information from data. The application ranges over several fields, including pharmacokinetics, such as determining clusters of drug absorption~\cite{tom:vin:car:17,gue:asmc:pmat:18}, weather prediction, e.g.~find patterns in hurricanes trajectories in order to better forecast the location of a hurricane landfall~\cite{Lee2007}, among others. Currently, with the massive introduction of electronic health records, there is a huge opportunity for mining temporal data with the objective of improving the prediction tasks in the biomedical domain~\cite{Zhao2017}.

The main objective of this work is the analysis of multivariate time series, namely, the imputation of missing values. Our approach is to use Dynamic Bayesian Networks, which can represent in a compact way relations between random variables~\cite{Pearl:1988:PRI:52121}. As such, this work follows the work of \cite{monteiro2015polynomial} and \cite{sousa2018polynomial} by extending their models. The proposed method is implemented and assessed in synthetic and real data.
The real datasets are from \textit{UCI Machine Learning Repository} \cite{Dua:2017} and \textit{UCR Time Series Classification Archive} \cite{UCRArchive}.

\subsection*{Related work}

In this paper we only consider the most used methods that support categorical time series, namely the last observation carried forward (LOCF), mode, EM and Amelia. The simplest method is the LOCF, where the missing value is simply imputed as the previous observed value. The Mode imputes the missing value with the most recurrent value. In expectation maximisation (EM) the  missing values are considered hidden variables, that are randomly initalized. Then, an EM procedure is performed to improve the distribution of these hidden variables in order to maximize the likelihood of the data. The idea is that the imputed value is the most probable value of the associated hidden variable. 

A more complex method is the so called Amelia~\cite{honaker2011amelia}.  This method employs a three-step approach to impute the missing values~\cite{rubin1976inference}, where the first step creates plausible values for the missing observations. These values are then used to impute the missing values. This process is repeated a number of times, from which it results from the creation of a number of "complete" datasets. In the second step, these datasets are then analysed using complete-data methods. Finally, the result of the analysis is then combined. 
In this work we use an R package implementation of \textit{Amelia II: A program for missing data} as a state-of-the-art imputation method.

\section{Background}

\subsection{Bayesian Networks}
\label{subsec:bn}
Let $X$ be a discrete random variable that takes values over the finite set $\mathcal{X}$. Moreover, let $\boldsymbol X = (X_1, ..., X_n)$ be a $n$-dimensional random vector, where each $X_i$ takes values in $\mathcal{X}_i=\{ x_{i1}, \dots, x_{ir_i}\}$, where $r_i$ the number of values that $X_i$ can take.
A \gls{BN} is composed of a \gls{DAG} that encodes a joint probability distribution over a set of random variables \cite{Pearl:1988:PRI:52121}.
\begin{definition}
	A $n$-dimensional \glsfirst{BN} is a triple $B = (\mathbf{X}, G, \Theta)$ where:
	\begin{itemize}
		\item  $\mathbf{X}$ is an $n$-dimensional finite random vector where each random variable $X_i$ ranges over by a finite domain $D_i$. Henceforward, it is denoted the joint domain by $D = \prod^{n}_{i = 1} D_i$.
		\item $G = (N,E)$ is a \gls{DAG} with nodes $N = \{X_1,\dots, X_n\}$ and edges $E$ representing direct dependencies between variables.
		\item $\Theta$ encodes  the parameters $\{\theta_{ijk}\}_{i \in \{1,\dots,n\}, j \in D_{\prod_{X_i}}, k \in D_i}$ of a network, given by:
		\begin{equation*}
			\theta_{ijk} = P_B\left(X_i=x_{ik} | \Pi _{X_i} = w_{ij}\right),
		\end{equation*}
		where $\Pi_{X_i}$ denotes the set of parents of the node $X_i$ in the \gls{DAG} $G$, $x_{ik}$ is the $k$-th values of $X_i$ and $w_{ij}$ is the $j$-th configuration of $\Pi_{X_i}$. Moreover, $q_i$ is the number of total parents configurations of $X_i$, $q_i = \prod_{X_j \in \Pi_{X_i}} r_j$.
	\end{itemize}
\end{definition}

Intuitively, the network structure of a \gls{DAG} G encodes conditional independence assumptions, where each random variable $X_i$ is only dependent of the descendants nodes and independent of its nondecendants, given its parents. These independence assumption are used within the chain rule to provide a factorized joint distribution over $\mathbf{X}$, defined as:
\begin{equation}
P_B(X_1,\dots, X_n) = \prod_{i = 1}^{n} P_B(X_i\mid\Pi_{X_i}).
\end{equation}
	
Learning a \gls{BN} $B=\left(\mathbf{X},G,\Theta\right)$ reduces to the problem of finding the network structure $G$ and the parameters $\Theta$ that best fits the data $D$. When data $D$ is complete, i.e, there are no missing values and hidden variables, being $D = \{\mathbf{y}_1,\dots,\mathbf{y}_N\}$ given by a set of $N$ i.i.d. instances, usually score-based learning is employed. Therein, a scoring criterion $\phi$ is used to measure how well a candidate network $B$ describes $D$. . However, since learning \gls{BN} is a NP-hard problem, the search space of possible solutions is explored by restring the solution space (tree-like \cite{friedman1997bayesian} or C$\kappa$G-like network structures \cite{sousa2018polynomial} or heuristic methods (greedy-hill climber \cite{heckerman1995learning}). In both cases, learning can be stated as an optimization problem:
\begin{equation}
\maximize_{B \in \mathcal{B}_n}\ \phi(B, D),
\end{equation}
where $\mathcal{B}_n$ corresponds to the set of \glspl{BN} with $n$ variables being searched.

\subsection{Dynamic Bayesian Networks}
\label{subsec:dbn}

The \gls{DBN} extends the representation of \gls{BN} to temporal processes. For simplicity, assume the discretization of time in time slices $\{0,\dots, T\}$. Let $\mathbf{X}[t]=(X_1[t], \dots,X_n[t])$ be a random vector that denotes the values of the set of random variables $\mathbf{X}$ over the time $t$. Moreover, let $\mathbf{X}[t_1:t_2]$ denote the set of random vectors $\mathbf{X}[t]$ over $t_1 \leq t \leq t_2$. Finally, let the joint probability distribution over the trajectory of a stochastic process from $\mathbf{X}[0]$ to $\mathbf{X}[T]$, $P(\mathbf{X}[0],\dots,\mathbf{X}[T])$, abbreviated with $P(\mathbf{X}[0:T])$, be defined as:
\begin{equation}
P(\mathbf{X}[0:T]) = P(\mathbf{X}[0])\prod_{t=1}^{T-1}P(\mathbf{X}[t+1]\mid\mathbf{X}[0:t]).
\end{equation}

One common approach to ease computation is to assume that the process is \textit{Markovian} in $\mathbf{X}$.

\begin{definition}
A stochastic process is said to satisfy the $m$-th order \textit{Markov} assumption if, for all $t\geq 0$:
\begin{equation}
	P(\mathbf{X}[t+1]\mid\mathbf{X}[0:t]) = P(\mathbf{X}[t+1]\mid\mathbf{X}[t-m+1:t]),
\end{equation}
where in this case $m$ is called the \textit{Markov} lag of the process.
\end{definition}

Another assumption that simplifies the computation of the joint probability distribution is to consider the process to be stationary. This assumption is usually adequate when number of time slices in the training data is small.

\begin{definition}
A Markovian process is said to be stationary if:
\begin{equation}
	P(\mathbf{X}[t+1]\ |\ \mathbf{X}[t]) \text{ is equal for all time slices } t \in \{0, ..., T-1\}.
\end{equation}
\end{definition}

Rigorously, a \gls{DBN} is composed of two networks: an initial \gls{BN} that encodes dependencies among variables in the initial state $\mathbf{X}[0]$, and a transition network that explain how dependencies flow forward in time. These dependencies include the intra-slice (from $\mathbf{X}[t]$ to $\mathbf{X}[t+1]$) and inter-slide (among $\mathbf{X}[t+1]$) dependencies.

\begin{definition}
A stationary first-order Markov \gls{DBN} consists of:
\begin{itemize}
\item  A prior network $B^0$, which specifies the distribution over the initial states $\mathbf{X}[0]$;
\item A transition network $B_{t}^{t+1}$ over the variables $\mathbf{X}[t] \cup \mathbf{X}[t + 1]$, for all $t$, that specifies the transition probability $P(\mathbf{X}[t+1]\mid \mathbf{X}[t])$.
\end{itemize}
\end{definition}

\begin{figure}[h]
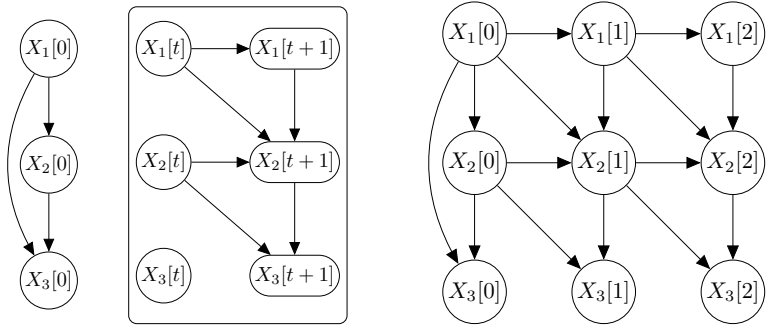

\centering
\begin{subfigure}[b]{0.45\linewidth}
	\centering
	 \resizebox{0.9\linewidth}{!}{
		\input{dbn}
	}
	\caption{Prior network (left) and  transition network (right) defining a \gls{DBN}.}
	\label{fig:dbn_norm}
\end{subfigure}
\begin{subfigure}[b]{0.45\linewidth}
	\centering
	 \resizebox{0.9\linewidth}{!}{
		\input{dbn_un}
	}
	\caption{The corresponding unrolled network for three time-slices.}
	\label{fig:dbn_un}
\end{subfigure}
\caption{Example of a stationary first-order \textit{Markov} \gls{DBN}.} 
\label{fig:dbn}
\end{figure}

An example of a \gls{DBN} along with its  unrolled network for three time-slices is depicted in \Cref{fig:dbn_un}.


In the case that all data is observed, there are three approaches for learning dynamic Bayesian networks. The first approach learn optimal \gls{DBN} using mutual information tests, but ignores intra-slice dependencies~\cite{vinh2011polynomial}. A polynomial-time algorithm that learns an 
To address both inter and intra-slice connections, a polynomial-time algorithm  was recently proposed~\cite{monteiro2015polynomial}. However, to keep the complexity low, the search space for the intra-slice connections is restricted to tree augmented networks, i.e, acyclic networks where each variable has only one parent from the same time slice, but can have a finite number of parents from the previous time slices. The resultant network is denoted by tDBN. More recently, the search space was extended exponentially~\cite{sou:car:18,sousa2018polynomial} to networks where the intra-slice network has in-degree at most $\kappa$ and is consistent with the \gls{BFS} order of the tDBN. The resultant network of this method is denoted by bcDBN.

\section{Structural EM}
\label{sec:learnfrommissing}

In many cases, the assumption that the training data are fully observed is simply unrealistic. Since this assumption is crucial for learning the structure and the parameters of a \gls{BN} some changes to the learning process need to be made.

When data $D$ is incomplete due to missing values or hidden variables, however, scoring functions are no longer decomposable. This shortcoming was addressed in~\cite{friedman1997bayesian},  where the \glsfirst{SEM} iterative method to learn a \gls{BN} with hidden variables and/or missing values, a brief description of this method can be seen in this section.

\subsection{Parameter Estimation}
The first learning task that will be considered is the parameter estimation task. As in the case of the complete data, the approach that will be used is the \gls{MLE}. So given a network structure $G$ and the form of the \glspl{CPD}, it is only necessary to compute the parameters $\Theta$ to define the distribution  $P(X\ |\ \Theta)$.  It is also given a data set $D$ that consists of $M$ partial instances of $X$, so it is needed to compute the values $\hat{\Theta}$ that maximize the log-likelihood function:

\begin{equation}
\hat{\Theta} = \argmax_{\Theta}\ \log L(\Theta:D).
\end{equation}


Unlike the complete data case, where sufficient statistics are collected for each  \gls{CPD} allowing to compute the parameters that maximize the likelihood, in the case of missing data there is no access to full sufficient statistics. In order to have access to them, one can take a simple approach of filling the missing values arbitrarily.  Some strategies to fill these missing values consist on choosing some default value or one according to some prior distribution. The problem with this approach is that the filled value will introduce bias in the learned parameters.

Another approach tries to solve two different problems at once, these problems being the learning of the parameters and imputation of the missing values.
To be noted that each of these tasks is very easy when the solution to the other is present.

The \gls{EM} algorithm starts by choosing some arbitrary starting point. This can be either a choice of parameters or some assignment to the missing values. Assuming that it begins with a parameter assignment, then the algorithms repeat two steps. The first step is to use the current parameters in order to complete the data, using probabilistic inference. The second step, consists in using the completed data as if it was observed and compute a new set of parameters. So, given a set of parameters $\theta^0$ and a partial instance, the posterior of all possible assignments to the missing value of that instance can be calculated. The \gls{EM} algorithm then uses this probabilistic completion of different data instances to estimate the expected value of the sufficient statistics. It is well known that  each iteration of this method increases the log-likelihood (see for instance~\cite{Koller:2009:PGM:1795555}), and moreover, that this process converges to a local maximum of the likelihood function.  

It is worthwhile to give a more detailed explanation of the \gls{EM} algorithm.
Assume the general \gls{BN} with table-\glspl{CPD} and an initial assignment for the parameters $\Theta^0$. Let $X$ be all the child variables, $W$ all the parent variables  and $o$ the dataset composed by $M$ data instances. The algorithm iterates over the following steps.

\paragraph{Expectation (E-Step)}In this step the \gls{ESS} are computed, this is done by using the current parameters $\Theta^t$.
\begin{itemize}
	\item For each family $X$, $W$ and for each data case $o[m]$, compute the joint probability $P(X,W\ |\ o[m], \Theta^t)$.
	\item Compute the \gls{ESS} for each $x$, $w$,
	\begin{equation}
	\bar{M}_{\theta^t}[x,w] = \sum_{m} P(x,w\ |\ o[m], \theta^t).
	\label{eq:ess}
	\end{equation}
\end{itemize}

\paragraph{Maximization (M-Step)} Given the \gls{ESS}, it performs maximum likelihood estimation, with respect to them, in order to compute a new set of parameters,
\begin{equation}
\theta^{t+1}_{x\ |\ w} = \frac{\bar{M}_{\theta_t}[x,w]}{\bar{M}_{\theta_t}[w]}.
\end{equation}
In \Cref{al:ess} and \Cref{al:em} the E-Step and for the Parameter EM are given.

\begin{algorithm}
	\caption{Compute the expected sufficient statistics}\label{al:ess}
	\begin{algorithmic}[1]
		\Procedure{Compute-ESS}{$G, \Theta, D$}
			\For{each $i = 1, \cdots, n$}\Comment{Initialization of data structures}
				\For{each $x_i, w_i\ \in Val(X_i, \Pi ^G_{X_i})$  }
					\State $\bar{M_t}[x_i, w_i] \gets 0$
				\EndFor
			\EndFor
			\For{each $m = 1, \cdots, M$}\Comment{Collect probabilities from all instances}
				\State Run inference on $ \langle G, \Theta \rangle$ using evidence $o[m]$
				\For{each $i = 1, \cdots, n$}
					\For{each $x_i, w_i\ \in Val(X_i, \Pi ^G_{X_i})$  }
							\State $\bar{M_t}[x_i, w_i] \gets \bar{M_t}[x_i, w_i] + P(x_i, w_i\ |\ o[m])$
					\EndFor	
				\EndFor
			\EndFor
			\State \textbf{return} $\{\bar{M_t}[x_i, w_i] : \forall i = 1, ..., n,\   \forall x_i, w_i \in Val(X_i, \Pi ^G_{X_i})\}$
		\EndProcedure
	\end{algorithmic}
\end{algorithm}

\begin{algorithm}
	\caption{Expectation-Maximization algorithm for Bayesian Network(using table-CPDs)}\label{al:em}
	\begin{algorithmic}[1]
		\Procedure{Expectation-Maximization}{$G, \Theta^0, D$}
			\For{each $t = 0, \cdots,$ until convergence}
				\State $\{\bar{M_t}[x_i, w_i]\} \gets$ Compute-ESS($G, \Theta^t, D$) \Comment{E Step}
				\For{each $i = 1, \cdots, n$}\Comment{M Step}
					\For{each $x_i, w_i\ \in Val(X_i, \Pi ^G_{X_i})$  }
						\State $\theta^{t+1}_{x_i|w_i} \gets \frac{\bar{M_t}[x_i, w_i]}{\bar{M_t}[w_i]}$
					\EndFor	
				\EndFor
			\EndFor
			\State \textbf{return} $\Theta^t$
		\EndProcedure
	\end{algorithmic}
\end{algorithm}

\subsection{Structure Learning}

The intuition behind Structural EM algorithm is the same that was applied to solve the problem of learning the parameters of a \gls{BN} when there is missing data. Like the parameter estimation, there is two main steps, the expectation, where a complete data set is generated, and a maximisation, where the network structure is learned. The main difference between the \glsfirst{SEM} and the parameter estimation is that the maximisation step, in the \gls{SEM}, besides learning the parameters, the network structure is also learned. Moreover, \cite{Koller:2009:PGM:1795555} state that by using the \gls{MDL} score it is guaranteed that, in each iteration, the learned structure is better than the one used in the previous iteration. From this statement, it results that the \gls{SEM} algorithm will monotonically improve the score.
The pseudo-code of the algorithm is given in \Cref{al:sem}.

\begin{algorithm}
	\caption{Structural EM algorithm for Bayesian Networks}\label{al:sem}
	\begin{algorithmic}[1]
		\Procedure{Structural-EM}{$G^0, \Theta^0, D$}
			\For{each $t = 0, \cdots,$ until convergence}
				\State ${\Theta^{t}}' \gets$ Expectation-Maximization($G^t, \Theta^t, D$) \Comment{Optional parameter learning step}
				\State $G^{t+1} \gets $ Structure-Learn($D_{G^t,{\Theta^{t}}'}^*$) \Comment{Run EM to generate the \gls{ESS} for $D_{G^t,{\Theta^{t}}'}^*$} 
				\State $\Theta^{t+1} \gets$ Estimate-Parameters($D_{G^t,{\Theta^{t}}'}^*, G^{t+1}$)
			\EndFor
			\State \textbf{return} $G^{t}, \Theta^{t}$
		\EndProcedure
	\end{algorithmic}
\end{algorithm}

\section{Proposed Method}
One common problem with multivariate time series is missing values. Mostly because many methods assume full data and are useless in this scenario. Thus, finding ways to work with missing values becomes crucial. One of the most used approaches to solve this problem is to drop the observations with missing values, however, when the dataset has few observations this approach can lead to enormous loss of information.

 Another approach that one can take is to impute the missing values. In this approach, the missing values are ``filled'' using some method, like an interpolation.  Since the focus of this thesis is the multivariate categorical time series, the most common methods for interpolation does not apply. So in order to impute the missing values, this work proposes a method that uses the \gls{SEM} algorithm, devised by \cite{friedman1998bayesian}, to learn the structure of the data with missing values. However, because the algorithm learns \glspl{BN}, it cannot model a time series, as such the algorithm was changed for the purpose of learning \glspl{DBN}.  As before, the search space is restricted to tDBN \cite{monteiro2015polynomial} and bcDBN \cite{sousa2018polynomial}. The \gls{SEM} algorithm can be divided with two big steps the parameter learning and the structure learning, and because the dataset has missing values one step cannot be learned without the other. As such, this algorithm starts by generating a \gls{DBN} randomly. Then the ``true'' parameters of the fixed network can be learned. This is done in an iterative process where first the \gls{ESS} are computed and then the new set of parameters are computed. This is done until convergence. With the parameters learned the algorithm then learns a new structure and repeats this process until the convergence criterion is met. 
 
Finally with the \gls{DBN} given by the \gls{SEM}, the imputation algorithm then generate again a new dataset without missing values, however instead of having all the possible combinations of values that could fill the missing values, this dataset fills the missing values with the combinations that maximizes the posterior probability.
In \cref{sec:learnfrommissing}  can be seen a description of the \gls{SEM} algorithm which will be the base to develop the imputation algorithm. 

\begin{algorithm}
	\caption{Missing values imputation via a DBN}\label{al:imp}
	\begin{algorithmic}[1]
		\Procedure{Imputation-DBN}{$D$}
			\State $G^0, \Theta^0 \leftarrow$  Generate a random \gls{DBN}
			\State $G, \Theta \leftarrow$  Structural-EM($G^0, \Theta^0, D$)
			\For{each observation in $D$}
				\For{each transition in the observation}
					\If{transition has missing values}
						\State Generate all possible combinations of values for the missing values
						\For{each new combination of values}
							\State Calculate the posterior probability
						\EndFor
						\State Select the generated combination that maximizes the posterior probability and impute the missing values 
					\EndIf
				\EndFor
				\State Add the observation to $D'$
			\EndFor
			\State \textbf{return} $D'$
		\EndProcedure
	\end{algorithmic}
\end{algorithm}

\section{Experimental Results}
\label{sec:results}

 \subsection{Simulated data}
 \label{subsec:impsim}
\begin{figure}[h]
 	\centering
 	\includegraphics[width=\linewidth]{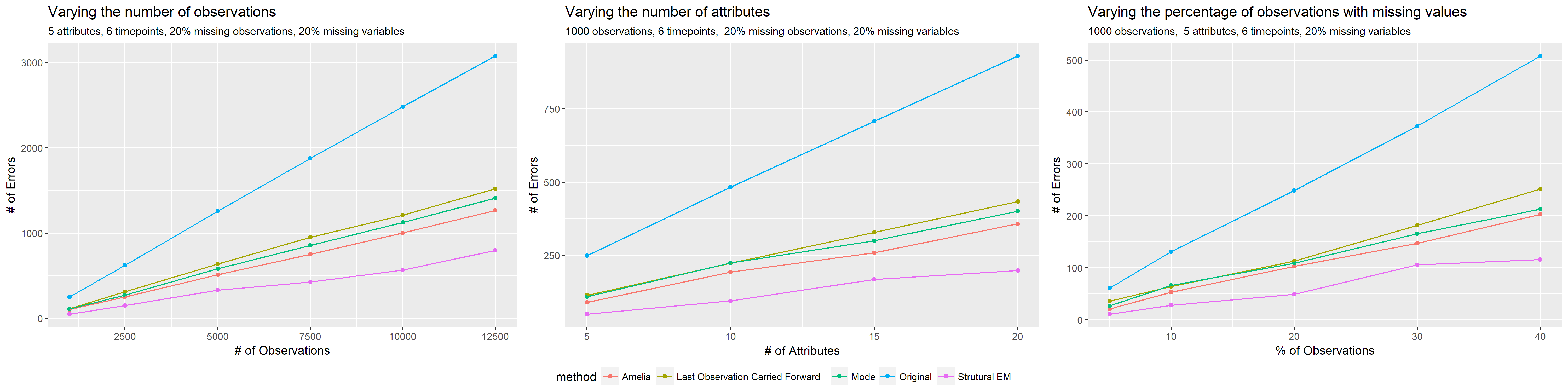}
 	\caption{Imputation errors for multiple datasets~(color online). }
 	\label{fig:pimperros}
 \end{figure}

  To assess the merits of the \gls{SEM} algorithm as an imputation method, multivariate time series were randomly generated using generated \glspl{DBN}. Then various datasets were generated, where the characteristics like the number of observations and the number of variables were changed. Finally, the missing values were generated with respect to two parameters, the first is the percentage of subjects with missing values and the second is the percentage of missing values corresponding to a subject. With the purpose of comparing this imputation method with state of the art methods, first these methods were used to impute the datasets with missing values, then the number of errors between the original dataset, without missing values, and the imputed datasets were counted. To facilitate the visualization, the results of this experiments are grouped in \Cref{fig:pimperros}.
 
 When analysing the results presented in \Cref{fig:pimperros} it can be concluded that the imputation done by \gls{SEM} algorithm has fewer errors when comparing with other imputation methods like \gls{LOCF}, Mode and the Amelia \cite{honaker2011amelia}. This result was expected because the data were generated using \glspl{DBN}, however, it important to highlight that despite the \gls{SEM} algorithm has fewer errors, it has errors.  One justification of this can be the fact that the imputation chosen by the algorithm is the one that maximizes the probability of the observation.

 \subsection{Benchmark data}
 \label{subsec:impreal}
  In order to evaluate the performance of the \gls{SEM} algorithm as an imputation method it was used 10 datasets from \textit{UCI Machine Learning Repository} \cite{Dua:2017} and \textit{UCR Time Series Classification Archive} \cite{UCRArchive}. Moreover, because the implementation of the \gls{SEM} algorithm only works with categorical time series and most of these datasets are composed by real-valued time series a discretization of these time series must be done. The discretization was done using the \textit{SAX} algorithm~\cite{lin2007experiencing}, it is important to note that it was used only an alphabet size of four, a maximum size of the time series of one hundred time steps and an independently discretization of each dimension of multivariate time series .
 
  With the resulting discretized datasets, in order to test the performance of the imputation methods, it was removed values. Once more, these  values were removed with respect to two parameters, the percentage of observation with missing values and the percentage of missing values per missing observation.
 
  Moreover, in order to use the \gls{SEM} algorithm some assumptions need to be made, these are the stationarity of the time series and the first-order \textit{Markov} assumption. Analysing the results, \Cref{fig:realerror}, it can be concluded that, in most datasets, the imputation done with \gls{SEM} algorithm has fewer errors than the other methods. 
 
 Given these results, a Wilcoxon signed ranks test was performed,  in order to compare the \gls{SEM} algorithm with others imputation methods. As such, the results were grouped by  methods and by the percentage of observations with missing values, \Cref{table:wil}. The use of a Wilcoxon signed ranks test is justified by the fact that is simple and a robust non-parametric test for statistical comparisons~\cite{demvsar2006statistical}.
 
  \begin{figure}[h]
 	\centering
 	\includegraphics[width=\linewidth]{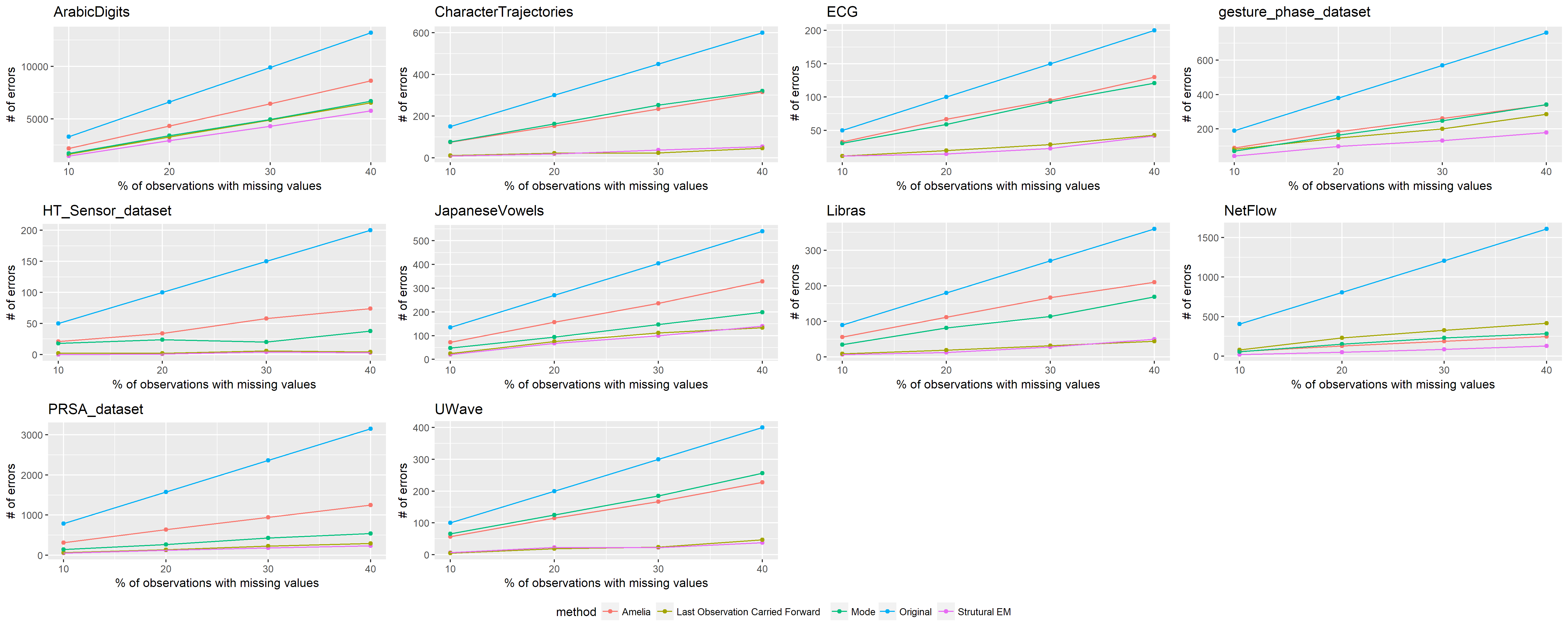}
 	\caption{Imputation errors for real datasets.}
 	\label{fig:realerror}
 \end{figure}
 
\begin{figure}[h]
 	\centering
 	\includegraphics[width=0.55\linewidth]{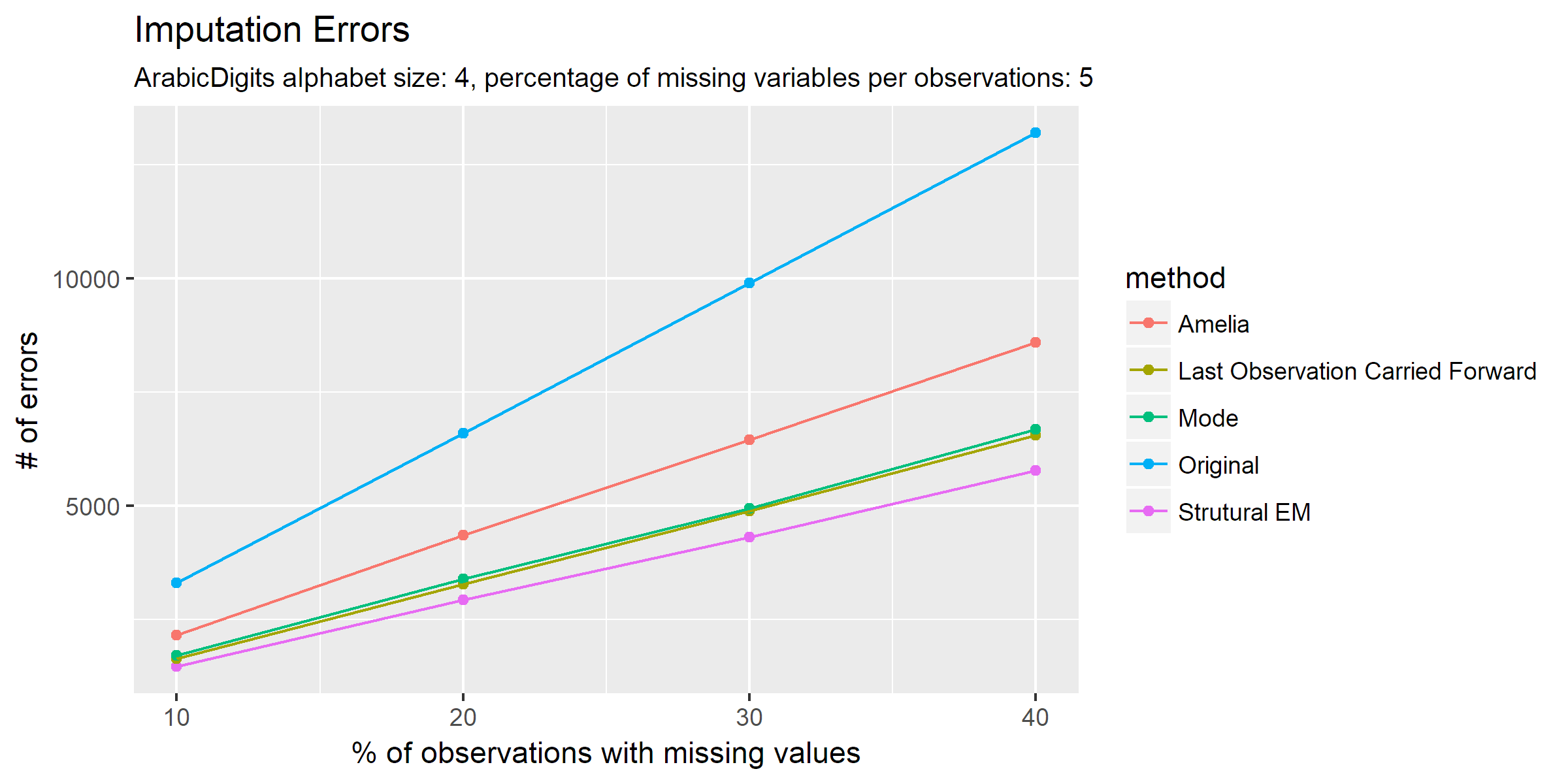}
 	\includegraphics[width=0.55\linewidth]{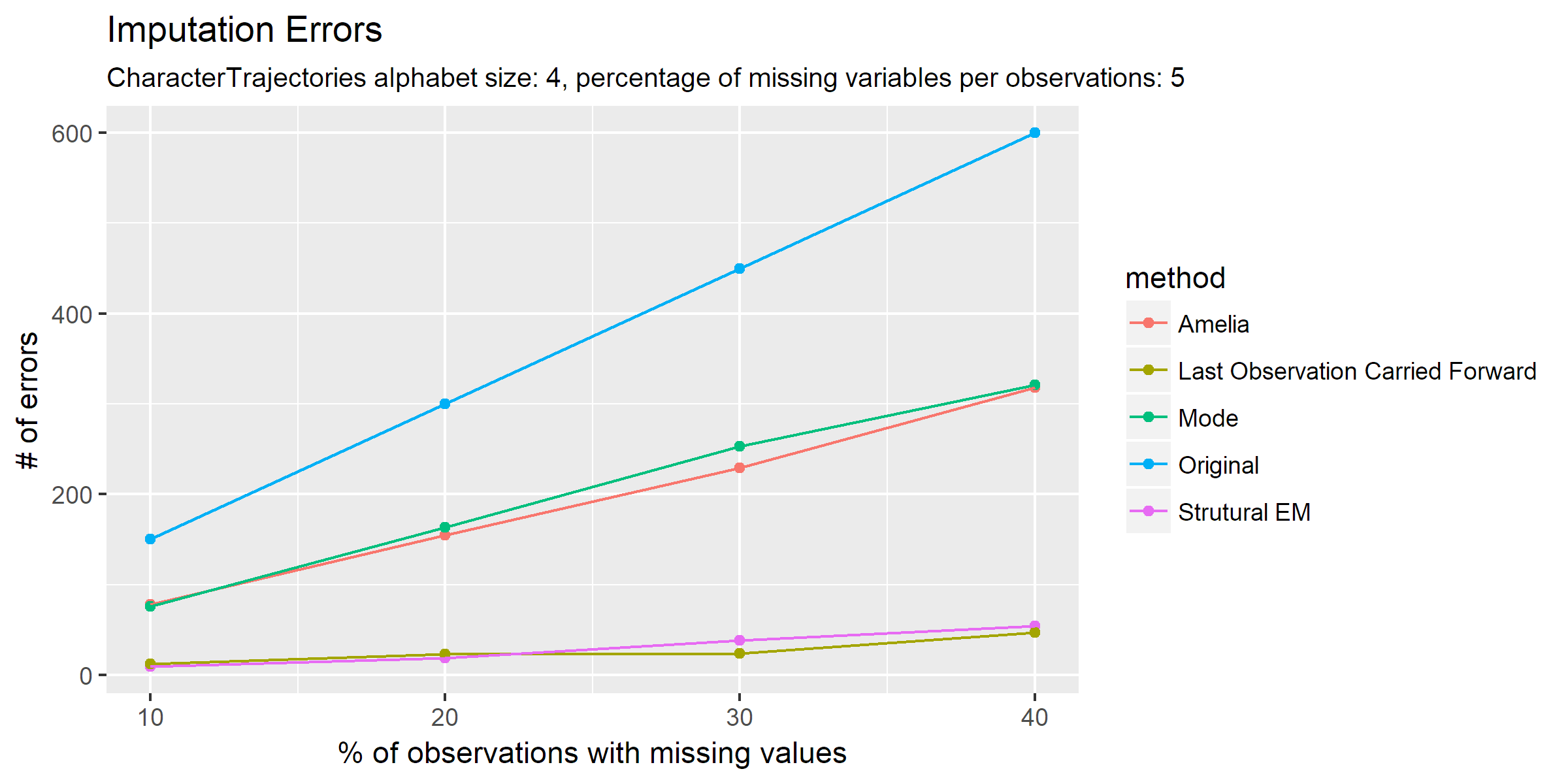}
 	\caption{Imputation errors for real datasets.}
 	\label{fig:realerror}
 \end{figure}
 
 \begin{table}[h]
 	\centering
 	\begingroup\catcode`"=9
 	\sisetup{round-mode=places,scientific-notation=true, round-precision=2}
 	\renewcommand{\arraystretch}{1.2}
 	\begin{filecontents*}{wilcoxon_test_new.csv}
",""Method"",""10"",""20"",""30"",""40"",""V""";;;;;;
"1,""LOCF"",""1.74E-02"",""1.25E-02"",""3.22E-02"",""1.26E-01"",""0""";;;;;;
"2,""Mode"",""5.89E-03"",""1.95E-03"",""1.95E-03"",""1.95E-03"",""0""";;;;;;
"3,""Amelia"",""1.95E-03"",""1.95E-03"",""1.95E-03"",""1.95E-03"",""0""";;;;;;
\end{filecontents*}
\csvreader[head to column names,
 	tabular = @{}lcccc@{}, 
 	table head =	\toprule \bfseries Method & 10 & 20 & 30 & 40 \\\midrule,
 	table foot = \hline]
 	{wilcoxon_test_new.csv}{}
 	{\csvcolii & \csvcoliii & \csvcoliv & \csvcolv & \csvcolvi}
 	\endgroup
 	\caption{Results from the Wilcoxon signed ranks test between \gls{SEM} algorithm and other methods.}
 	\label{table:wil}
 \end{table} 
 
 When analysing results from \Cref{table:wil} it is easy to note that almost all p-values have a value below 0.05, which indicates that the null-hypothesis, that the two algorithms perform equally well, is discarded. However, it is important to note that the imputation errors of \gls{LOCF} are similar to the imputation errors of the \gls{SEM} algorithm which is a strange result. One explanation for this result can be the number of symbols used to discretize the time series, because since 4 symbols were used the discretized time series may not vary that much over time, which leads to a lower imputation error when using the \gls{LOCF}.

\end{document}